\documentclass{article}
\usepackage{spconf,amsmath,graphicx,hyperref}
\usepackage{booktabs}
\usepackage{multirow}
\usepackage{amsmath}  
\usepackage{xcolor}
\usepackage{amssymb}
\usepackage{caption}

\title{Judge Before Answer: Can MLLM Discern the False Premise in Question?}
\name{Jidong Li$^1$, Lingyong Fang$^1$, Haodong Zhao$^1$, Sufeng Duan$^1$, Gongshen Liu$^{1,2}$ }
\address{$^1$School of Computer Science, Shanghai Jiao Tong University \\ $^2$Inner Mongolia Research Institute, Shanghai Jiao Tong University
}
\begin{document}
\ninept
\maketitle
\begin{abstract}

Multimodal large language models (MLLMs) have witnessed astonishing advancements in recent years. Despite these successes, MLLMs remain vulnerable to flase premise problems. However, existing benchmarks targeting this issue are limited in scope: they often lack fine-grained categorization, exhibit insufficient coverage, and thus fail to provide a rigorous evaluation of the ability of models to recognize false premises.

To bridge this gap, we introduce a fully automated pipeline for constructing a comprehensive benchmark of false premise questions. Our method systematically categorizes the premises into three main types and thirteen subtypes according to the abilities required to identify the premises, resulting in the JBA dataset.Results show current MLLMs still struggle with false premise recognition. Building upon this benchmark, we further propose a recognition enhancement framework tailored to strengthen the robustness of MLLMs to detect false premises. Extensive experiments demonstrate that models trained with our framework achieve significant improvements in false premise recognition.
Our code is publicly available at \href{https://github.com/JidongLi-hub/JudgeBeforeAnswer}{https://github.com/JidongLi-hub/JudgeBeforeAnswer}.
\end{abstract}
\section{Introduction}
\label{sec:intro}


The evolution of Multimodal Large Language Models (MLLMs)~\cite{radford2021learning} has marked a significant milestone in artificial intelligence, with models exhibiting remarkable capabilities in handling complex tasks that integrate vision and language. However, a paradox has emerged: this exponential growth in capability has not been matched by a corresponding increase in reliability and trustworthiness. A critical manifestation of this brittleness is the ``false premise problem'', which arises when a model is presented with a question containing factually incorrect or non-sensical presuppositions. An ideal model would identify and refute the invalid premise. Instead, current MLLMs often accept the premise by default, proceeding with unnecessary and flawed reasoning that culminates in a confidently delivered, yet entirely unreliable answer~\cite{dont}.

Research on false premise problems initially started with single-modal language models. For instance, benchmarks like \textit{FalseQA}~\cite{wont} revealed that while LLMs possess some capacity to recognize false premises, they often fail to refute them explicitly. Similarly, work in mathematical reasoning~\cite{Li2024b}, such as \textit{MathClean}~\cite{Liang2025}, has exposed the limitations of even powerful models in detecting defects within problems. \textit{MathQ-Verify}~\cite{Shen2025} to filter ill-posed mathematical questions, thus improving the verification performance. 
In addition, \textit{DecoPrompt}~\cite{decoprompt} introduced a prompting strategy that leverages LLM to ``decode'' false premises in order to mitigate hallucination-prone prompting algorithms.
As research pivots to the more complex multimodal space, the problem is amplified. A work introduces the concept of unresolvable detection in MLLMs\cite{unsolvable}.
While early multimodal benchmarks like \textit{MAD-Bench}~\cite{howeasy} and \textit{ISEval}~\cite{mllmdataset} have initiated the investigation, they expose the clear need for more sophisticated evaluation tools. There are also methods that introduce additional steps, enabling the model to segment the image to verify the existence of premises rather than making direct judgments\cite{see}. These existing datasets are often limited in scale, with MAD-Bench containing fewer than 1,000 samples, or they lack the fine-grained categorization necessary for deep diagnostics.


Beyond dataset construction, another line of research has focused on enhancing models’ ability to recognize or explicitly reject questions containing false premises. 
For instance, some researchers attempted to strengthen the ability of models to respond with ``I don’t know'' when facing ambiguous queries~\cite{visually}. 
Similarly, works such as \textit{FalseQA}~\cite{wont} and \textit{ISEval}~\cite{mllmdataset} highlight that while current LLMs and MLLMs have partial capabilities in recognizing false premises, their success is highly dependent on whether they are explicitly guided or equipped with refutation mechanisms.

This landscape reveals two fundamental gaps in current research. First, there is an \textbf{Evaluation Gap}: a lack of a fine-grained, comprehensive, and large-scale dataset to precisely measure the ability of MLLMs to recognize false premises. Existing datasets are often limited in their coverage and scope. Second, there is a
\textbf{Methodological Gap}: a need for a targeted training framework to enhance a model's ability to scrutinize and reject false premises. While some studies have focused on prompting models to respond with ``I don't know'' to ambiguous queries, a dedicated approach to instill this skill is missing~\cite{visually}. 

\begin{figure*}[htbp]
    \centering
    \includegraphics[width=0.7\textwidth]{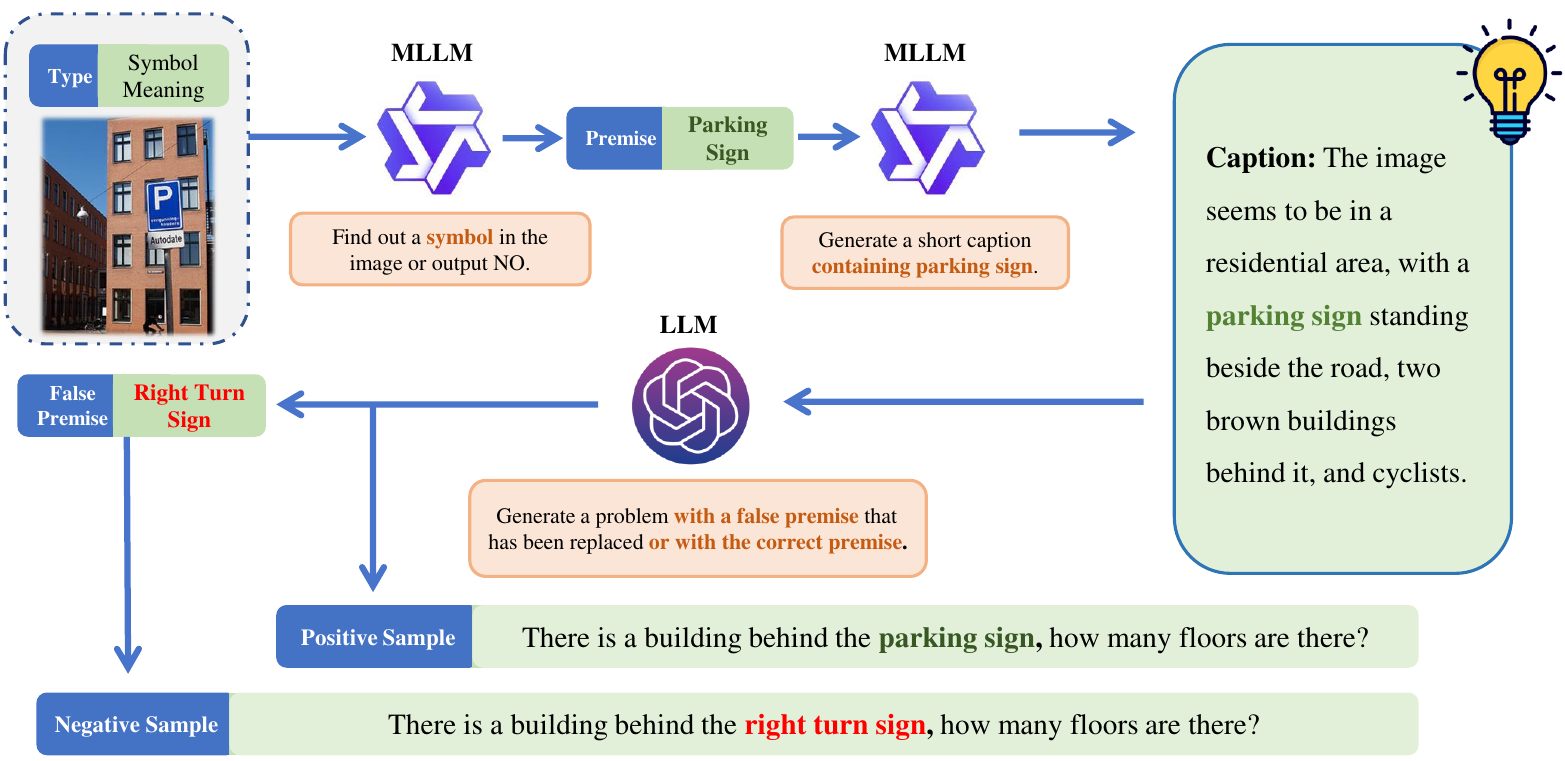}
    \captionsetup{skip=5pt}
    \caption{The fully automated pipeline for constructing our false premise dataset, consisting of three stages: Visual Premise Extraction, Premise-Aware Captioning, and Target Question Generation. First, an MLLM extracts a premise of a specific type from an input image. Next, it generates a concise caption that must include the extracted premise. Finally, an LLM produces a false premise question or a true premise question by embedding either a false premise obtained through replacement nor the correct premise into declarative forms.}
    \label{fig:pipeline}
    \vskip -0.2in
\end{figure*}
This paper addresses these gaps with a synergistic, two-part solution. First, we introduce JBA, a comprehensive dataset for false premise questions, built using a fully automated construction pipeline. This method categorizes premises into a hierarchical taxonomy of three levels and thirteen subtypes, creating a systematic dataset while significantly reducing annotation costs. Second, we propose \textbf{JBA-GRPO}, a reinforcement learning framework designed to enhance an MLLM's ability to recognize and explicitly address such errors. This framework leverages the training data conveniently generated by the automated pipeline, creating a tailored and effective training regimen. Extensive experiments show that representative MLLMs still perform poorly on this task, but a model trained with the JBA-GRPO framework achieves substantially better performance, validating the effectiveness of this approach. The main contributions are as follows:

$\bullet$ We introduce JBA, a comprehensive benchmark for evaluating false premise detection in MLLMs, created through an innovative automated pipeline. It features a hierarchical taxonomy of 13 subtypes across three cognitive levels for detailed analysis.

$\bullet$ We propose JBA-GRPO, a reinforcement learning framework to enhance MLLM in identifying and rejecting false premises, incorporating a novel ``reasoning reward'' to ensure logical coherence.

$\bullet$ Experiments reveal systemic weaknesses in existing MLLM and demonstrate the superiority of the JBA-GRPO trained model, establishing a new standard for reliable multimodal reasoning.\looseness=-1






\section{JBA Dataset}
\label{sec:dataset}

\subsection{Construction Pipeline}\label{sec:pipeline}
In this section, we propose a fully automated pipeline for generating false premise question datasets. By decomposing the task into modular steps and leveraging the advanced text and image understanding capabilities of state-of-the-art LLMs and MLLMs, our approach enables the efficient construction of large-scale, diverse, and fine-grained false premise datasets.

 Fig.~\ref{fig:pipeline} illustrates the fully automated pipeline for constructing our false premise dataset, which consists of three main stages. Let $I$ denote an input image, $t$ a specific premise type, $p$ the extracted premise content, $c$ the premise-aware caption, and $q$ the generated question. The pipeline is formally described as follows:

\subsubsection{Visual Premise Extraction}
Given an image $I$ and a premise type $t$ (premise types are one of the key factors ensuring dataset comprehensiveness and are described in detail in Section~\ref{sec:jba}; in our running example, $t$ is symbol meaning, an MLLM is prompted to detect whether $I$ contains content corresponding to $t$. 
\[
p = \text{MLLM\_Extract}(I, t).
\]  
 If the premise is present, $p$ is output and forwarded to the next stage.  
 If the premise is absent, the model outputs \texttt{NO}, indicating a failed match, and the next candidate image or premise type is screened.  

\subsubsection{Premise-Aware Captioning}
Based on the extracted premise $p$, the MLLM is prompted to generate a concise caption $c$ for the image $I$ that explicitly includes $p$ or treats it as the central focus of the description:  
\[
c = \text{MLLM\_Caption}(I, p).
\]  
The goal is not to produce a detailed description of the entire image, but to provide a few sentences that capture the premise. This compression of visual information simplifies and improves the efficiency of subsequent premise replacement in the next stage.  

\subsubsection{Target Question Generation}
Given the premise-aware caption $c$ and the original premise $p$, another LLM generates both positive and negative question samples $q$:\looseness=-1
\[
q = \text{LLM\_Question}(c, p).
\]

 \textbf{Negative samples:} Replace the original premise $p$ with an incorrect or contradictory (but never identical) premise $p^{\prime}$. The LLM then generates a question that incorporates $p^{\prime}$ declaratively while querying other aspects of the caption. Formally:\looseness=-1
    \[
    p^{\prime} \sim \text{IncorrectPremise}(p), \quad q_\text{neg} = \text{LLM\_Generate}(c, p^{\prime}).
    \]    

 \textbf{Positive samples:} No replacement is performed, and the LLM generates a question that includes the original correct premise $p$ declaratively, without directly asking about $p$:  
    \[
    q_\text{pos} = \text{LLM\_Generate}(c, p).
    \]  
    
Carefully designed few-shot~\cite{fewshot} examples are incorporated in each step above to ensure the reliability of premise extraction and replacement as well as the quality of the generated questions.

\begin{figure}[htbp]
    \centering
    \includegraphics[width=0.5\textwidth]{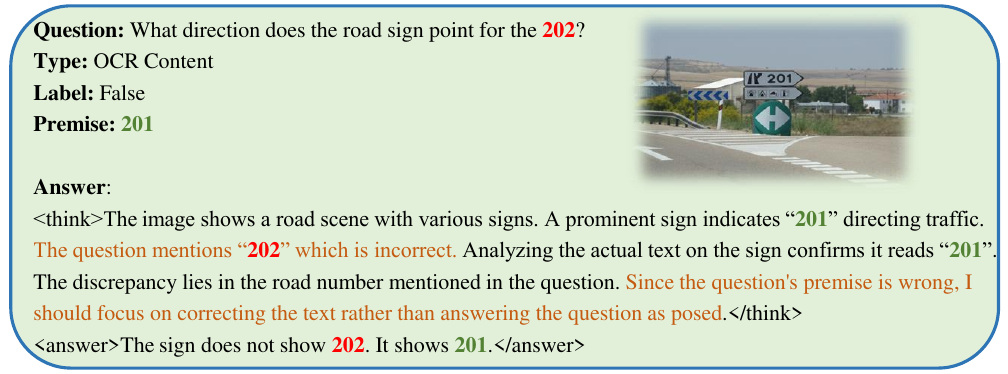}
    \caption{An example data used in the JBA-GRPO training.
}
    \label{fig:sample}
    \vskip -0.2in
\end{figure}

 Furthermore, we can generate specific answers for the questions, which serve as training data as described in Section\ref{sec:grpo}. Fig.~\ref{fig:sample} illustrates an example of such training data, where the answer analyzes the false premise and explicitly points out the error.

\subsection{JBA Dataset}
\label{sec:jba}

\begin{figure}[htbp]
    \centering
    \includegraphics[width=0.33\textwidth]{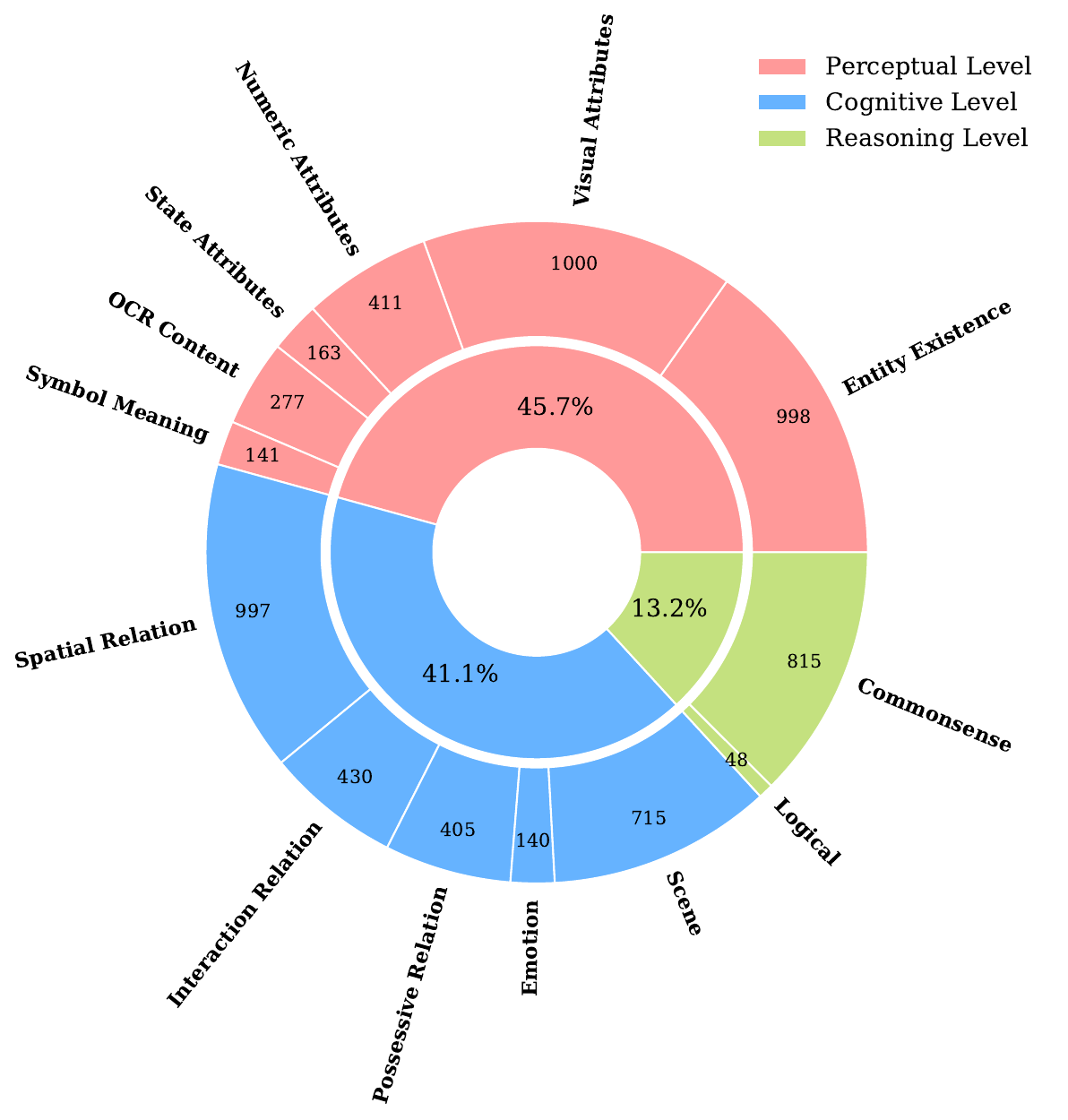}
    \caption{JBA Dataset Category Distribution}
    \label{fig:dataset}
    \vskip -0.1in
\end{figure}

Based on the pipeline in Section~\ref{sec:pipeline}, we establish a hierarchical taxonomy of false premises for systematic evaluation. As shown in Fig.~\ref{fig:dataset}, the taxonomy consists of three levels—Perceptual, Cognitive, and Reasoning—covering thirteen distinct categories, enabling structured analysis of model performance from basic perception to advanced reasoning.All images we use are sourced from the \textit{Visual Genome} dataset~\cite{vg}. We use Qwen2.5-VL-72B-Instruct~\cite{qwen2.5} as our MLLM and Qwen3-32B~\cite{qwen3} as our LLM.The following are the category descriptions.

\textbf{Perceptual-Level Premises.} This level focuses on directly observable aspects of images and includes six categories: \textit{Entity Existence}, \textit{Visual Attributes}, \textit{Numerical Attributes}, \textit{State Attributes}, \textit{Textual Content}, and \textit{Symbolic Meaning}. These categories primarily assess model ability to perceive and describe visual content accurately.

\textbf{Cognitive-Level Premises.} This level requires understanding higher-order relations and contextual information. It comprises five categories: \textit{Spatial Relations}, \textit{Interaction Relations}, \textit{Part–Whole Relations}, \textit{Emotional State}, and \textit{Scene}. Collectively, they evaluate whether models can integrate and reason over structured relationships within the image.

\textbf{Reasoning-Level Premises.} The highest level demands inference beyond the visible content and includes two categories: \textit{Logical Relations} and \textit{Commonsense Knowledge}. This level tests a model's ability to combine perception with abstract reasoning to interpret complex or implicit aspects of the scene.

\begin{table*}[ht]
\centering
\small
\resizebox{\textwidth}{!}{
\begin{tabular}{lccccccccc}
\toprule
\multirow{2}{*}{\textbf{Models}} 
  & \multicolumn{3}{c}{\textbf{Perceptual Level}} 
  & \multicolumn{3}{c}{\textbf{Cognitive Level}} 
  & \multicolumn{3}{c}{\textbf{Reasoning Level}} \\
\cmidrule(lr){2-4} \cmidrule(lr){5-7} \cmidrule(lr){8-10}
 & \textbf{FPC} & \textbf{FPDP} & \textbf{TPIR} & \textbf{FPC} & \textbf{FPDP} & \textbf{TPIR} 
 & \textbf{FPC} & \textbf{FPDP} & \textbf{TPIR}  \\
\midrule

LLaVA-OneVision&58.7±0.05 &78.0±0.12&55.4±0.06
                             &61.6±0.06  &81.3±0.11&57.1±0.07
                               &77.3±0.09 &92.5±0.09&69.7±0.12\\
LLaVA-v1.5-7B &60.1±0.06&67.6±0.11&57.3±0.07&67.1±0.06&72.3±0.09
                &63.9±0.08&82.2±0.08&84.6±0.11&79.9±0.12\\
InternVL3-8B &74.8±0.05&86.2±0.07&69.0±0.06
               & 65.7±0.06&77.1±0.09&61.0±0.07
                &81.7±0.08&\textbf{93.0±0.09}&74.7±0.12\\
Qwen2.5-VL-7B-Instruct &77.7±0.05&84.6±0.06&73.2±0.07
                    &71.2±0.05&81.6±0.08&65.9±0.07
                        &84.3±0.08&87.7±0.1&81.2±0.11\\
                        
\midrule
JBA(Ours) &  \textbf{81.1±0.05} &  \textbf{86.6±0.06}  & \textbf{77.2±0.06}  &
                        \textbf{75.5±0.05}   &  \textbf{84.5±0.07}  &   \textbf{70.2±0.07}  &
                      \textbf{ 86.8±0.07 }    &   89.2±0.09 &  \textbf{84.5±0.11} \\
\bottomrule
\end{tabular}
}
\caption{Evaluation results across different levels. The best results are highlighted in \textbf{bold}. The same applies to the subsequent tables.
}
\label{tab:1}
\vskip -0.2in
\end{table*}

\section{JBA-GRPO}
\label{sec:grpo}

\subsection{Preliminary}

Group Relative Policy Optimization (GRPO) is a reinforcement learning strategy that enhances the deliberative capabilities of large language models for accuracy-focused tasks like mathematics, as implemented in the Deepseek-R1 model~\cite{grpo,guo2025deepseek}.

GRPO's objective (Eq.~\ref{eq:grpo_alt}) is to maximize a composite function that balances reward with training stability. It optimizes for the expected aggregate reward $R(o)$ while a Kullback-Leibler (KL)~\cite{kl} divergence term, scaled by $\beta$, regularizes the policy $\pi_\theta$ by penalizing deviation from a reference policy $\pi_\mathrm{ref}$. The expectation is taken over outputs sampled from a previous policy, $\pi_{\theta_{\mathrm{old}}}$.
\begin{equation}
\label{eq:grpo_alt}
    \max_{\pi_\theta} \mathbb{E}_{o\sim \pi_{\theta_{\mathrm{old}}}(p)} [
        R(o) - 
        \beta \mathrm{D}_\mathrm{KL}(\pi_\theta \| \pi_\mathrm{ref})
    ].
\end{equation}
The aggregate reward $R(o)$ is a weighted average over a group of $G$ candidate outputs (Eq. \ref{eq:ro_alt}). For each candidate, its score from a reward function $r(\cdot)$ is standardized (centered and scaled by the group's mean and standard deviation). This standardized score is then weighted by an importance sampling ratio, $\pi_\theta(o_i)/\pi_{\theta_{\mathrm{old}}}(o_i)$, to correct for sampling from an older policy.
\begin{equation}
\label{eq:ro_alt}
    R(o)=\sum_{i=1}^G\frac{\pi_\theta (o_i)}{\pi_{\theta_{\mathrm{old}}}(o_i)} \cdot
    \frac{r(o_i)-\mathrm{mean}(\{r(o_i)\}_{i=1}^G)}{\mathrm{std}(\{r(o_i)\}_{i=1}^G)}.
\end{equation}

\subsection{Reward Design}

In addition to the commonly used format reward and answer reward, we have also designed a reasoning reward.

Although the first two rewards enforce structural correctness and final answer accuracy, they are insufficient to ensure the logical integrity of the reasoning process itself. A model could potentially arrive at the correct answer through a flawed or coincidental chain of thought, a phenomenon detrimental to robust learning. For instance, within the \texttt{<think>} block, the model might present a line of reasoning that is logically inconsistent or even directly contradicts its own final conclusion.

To mitigate this, we employ a large language model (LLM) as an evaluator to assess the quality of the reasoning articulated in the \texttt{<think>} tag. This is accomplished using a meticulously crafted prompt that provides the evaluator LLM with detailed criteria. The LLM evaluates aspects such as logical coherence, relevance to the question, and whether the step-by-step reasoning genuinely supports the final answer. High-quality, coherent reasoning receives a positive reward, whereas flawed or irrelevant reasoning incurs a penalty.

\subsection{Training Strategy}

We trained Qwen2.5-VL-7B using a two-stage process on our JBA dataset. First, a Supervised Fine-Tuning (SFT) phase on the SFT subset aligned the model with our required reasoning structure and established baseline vision-language skills. Building on this SFT-tuned checkpoint, we then performed a Reinforcement Learning (RL) optimization phase. This second stage used the RL subset and three custom reward functions to refine the model's advanced reasoning abilities.

\section{Experiment}
\label{sec:Experiment}
\subsection{Evaluated Models}
To evaluate the capability of existing MLLMs in recognizing false premise questions, we conduct experiments on several representative models, including \textit{InternVL3}~\cite{internvl3}, \textit{Qwen2.5-VL}~\cite{qwenvl}, \textit{LLaVA-v1.5-7B}~\cite{llava}, and \textit{LLaVA-OneVision}~\cite{llavaone}. In addition, we further apply the reinforcement learning method proposed in this work to fine-tune \textit{Qwen2.5-VL-7B-Instruct}~\cite{qwenvl}, with the goal of enhancing its ability to identify false premises more effectively.

\subsection{Evaluation Metrics}

To evaluate the capability of different models in recognizing false premise questions on the JBA dataset, we introduce three metrics. 

Let $N$ denote the total number of questions in the dataset. For each question $i \in \{1, \ldots, N\}$, we define binary variables $y_i^\text{FP}, y_i^\text{TP} \in \{0,1\}$ to indicate the ground-truth labels of False Premise and True Premise, respectively. Similarly, $\hat{y}_i^\text{FP}, \hat{y}_i^\text{TP} \in \{0,1\}$ denote the corresponding model predictions. These notations allow us to formally specify the evaluation metrics as follows.
\begin{equation}
   \text{FPC} = \frac{1}{N} \sum_{i=1}^{N} (\hat{y}_i^\text{FP}+\hat{y}_i^\text{TP} ),
   \label{eq:fpc}
\end{equation}
\begin{equation}
   \text{FPDP} = \frac{\sum_{i=1}^{N} y_i^\text{FP} \cdot \hat{y}_i^\text{FP}}{\sum_{i=1}^{N} \hat{y}_i^\text{FP}},
   \label{eq:fpdp}
\end{equation}
\begin{equation}
   \text{TPIR} = \frac{\sum_{i=1}^{N} y_i^\text{TP} \cdot \hat{y}_i^\text{TP}}{\sum_{i=1}^{N} \hat{y}_i^\text{TP}}.
   \label{eq:tpir}
\end{equation}



\textbf{False Premise Coverage (FPC)} measures the proportion of all questions for which the model correctly identifies a false premise, reflecting its overall detection ability.  

\textbf{False Premise Detection Precision (FPDP)} is the proportion of correctly identified false premise questions among those predicted as false premise, indicating detection accuracy.  

\textbf{True Premise Identification Rate (TPIR)} measures the proportion of correctly recognized true premise questions among those predicted as true premise, reflecting reliability in identifying non-false premises.

\subsection{Main Results}


Table~\ref{tab:2} shows model performance on whole JBA dataset. Most baselines score low on TPIR, indicating limited ability to detect rare or challenging false premises, while higher FPDP scores suggest they can generally reject false premises accurately. FPC reflects overall recognition across all samples, highlighting substantial room for improvement to ensure more reliable responses.

\begin{table}[ht]
\centering
\small
\begin{tabular}{lccc}
\toprule
\textbf{Models}
  & \textbf{FPC ($\uparrow$)}
  & \textbf{FPDP ($\uparrow$)}
  & \textbf{TPIR ($\uparrow$)} \\
\midrule
LLaVA-OneVision &62.4±0.04&82.7±0.08 & 57.6±0.05\\
LLaVA-v1.5-7B &65.9±0.04&72.9±0.07& 62.2±0.05\\
InternVL3-8B &72.0±0.04&84.0±0.06&66.3±0.05\\
Qwen2.5-VL-7B-Instruct &75.9±0.04& 84.0±0.05 & 70.9±0.05\\

\midrule
JBA(Ours) & \textbf{79.5±0.03} & \textbf{86.2±0.05} & \textbf{74.9±0.05} \\

\bottomrule
\end{tabular}
\caption{Evaluation results on whole JBA dataset. }
\label{tab:2}
\vskip -0.1in
\end{table}

Table~\ref{tab:1} reports the model performance on the three levels of the JBA dataset. 
At the \textbf{perceptual level}, most models achieve relatively high scores on FPDP, indicating that they can stably exclude explicit false premises. 
However, their TPIR scores are generally low, suggesting that they tend to overlook complex or ambiguous perceptual errors. 
At the \textbf{cognitive level}, InternVL3-7B and Qwen2.5-VL-7B show a decline across all metrics, whereas LLaVA-OneVision and LLaVA-v1.5-7B exhibit improvements in each metric. 
At the \textbf{reasoning level}, all models perform better than at the other two levels, with InternVL3 achieving an FPDP of 93\%, demonstrating a strong ability to identify reasoning-related false premises. 
Although the difficulty of recognizing different types of false premise questions is inherently fixed, the results indicate that different models still exhibit preferences for certain types of premises.



Overall, the \textit{Qwen2.5-VL-7B-Instruct} model exhibits significantly stronger capabilities in recognizing false premises compared to other baseline models. Moreover, our proposed JBA model consistently outperforms these baselines across a range of evaluation metrics. The notable performance improvement of JBA can be attributed to two key design principles. 

First, we introduced a structured reasoning mechanism by incorporating a \texttt{<think>} tag into the model's inference pipeline. This tag enforces an explicit pre-analysis phase, during which the model systematically examines both the visual content and the textual premise of the input. Through this process, the model is able to deconstruct and compare the components of the question, thereby identifying potential discrepancies with greater precision.

Second, we employed reinforcement learning (RL) to further refine and generalize the model's reasoning ability. By rewarding the model for accurately detecting premise errors through its structured thought process, the RL framework enhances the robustness and adaptability of the model's inferential capabilities, particularly when dealing with novel or unseen examples. 

As a result, the superior performance of the JBA model in most comparative evaluations validates the effectiveness of integrating explicit analytical reasoning with reinforcement learning for the task of robust premise-error recognition.


\section{CONCLUSIONS}
\label{sec:conclusion}
This paper addresses the shortcomings of current multimodal false premise datasets, which are often limited in scope and costly to annotate. We introduce a fully automated pipeline using task decomposition and prompt design to construct JBA, a comprehensive dataset categorizing false premise questions into three levels and 13 subtypes. Evaluations on JBA reveal that while MLLMs show capability in recognizing false premises, their performance is limited, indicating room for improvement. To enhance robustness, we develop a tailored reinforcement framework for false premise recognition. Our models consistently outperform baselines, validating the efficacy and establishing JBA as a valuable dataset. These findings highlight the need for specialized training to enhance MLLMs' ability to detect and manage false premises across diverse scenarios.\looseness=-1

\newpage




\bibliographystyle{IEEEbib}
\bibliography{strings,refs}

\end{document}